\documentclass[conference]{IEEEtran}

\IEEEoverridecommandlockouts
\usepackage{cite}
\usepackage{amsmath,amssymb,amsfonts}
\usepackage{algorithmic}
\usepackage{graphicx}
\usepackage{subcaption}

\usepackage{float}
\usepackage{xcolor}
\usepackage{textcomp}
\usepackage{xcolor}

\usepackage{tikz}
\usetikzlibrary{arrows.meta,angles,quotes,calc}
\usepackage{tikz}
\usetikzlibrary{arrows.meta,angles,quotes,calc,decorations.pathreplacing}
\def\BibTeX{{\rm B\kern-.05em{\sc i\kern-.025em b}\kern-.08em T\kern-.1667em\lower.7ex\hbox{E}\kern-.125emX}}
\begin{document}


\title{Dynamic Modeling and Attitude Control of a Reaction-Wheel-Based Low-Gravity Bipedal Hopper}


\author{
\IEEEauthorblockN{
Shriram Hari$^{1}$, M Venkata Sai Nikhil, R Prasanth Kumar$^{2}$
}
\IEEEauthorblockA{
\textit{Department of Mechanical and Aerospace Engineering} \\
\textit{Indian Institute of Technology Hyderabad, Sangareddy, India} \\
$^{1}$\textit{(me22btech11023@iith.ac.in)}, 
$^{2}$\textit{(rpkumar@mae.iith.ac.in)}
}
}
\maketitle

\begin{abstract}
Planetary bodies characterized by low gravitational acceleration, including the Moon and near-Earth asteroids, impose unique locomotion constraints owing to diminished contact forces and extended airborne intervals. Among traversal strategies, hopping locomotion stands out for its energy efficiency; however, it suffers from mid-flight attitude instability triggered by asymmetric thrust generation and uneven terrain. This paper introduces an underactuated bipedal hopping robot that utilizes an internal reaction wheel to regulate its body posture during the ballistic flight phase. Modeled as a gyrostat, this design facilitates the analysis of the dynamic coupling between torso rotation and reaction wheel momentum. The locomotion cycle consists of three distinct phases: a leg-driven propulsive jump, mid-air attitude adjustment via an active momentum exchange controller, and a shock-absorbing landing. A reduced-order model is formulated to describe the critical coupling between torso rotation and reaction wheel spin. The framework is evaluated within a MuJoCo-based simulation configured for Moon-gravity conditions ($g = 1.625 \text{ m/s}^2$). Experimental comparisons confirm the effectiveness of the approach: activation of the reaction wheel controller reduces peak mid-air angular deviation by over 65\% and successfully bounds landing attitude errors to within $3.5^\circ$ at touchdown. Furthermore, actuator saturation is reduced per hop cycle, ensures sufficient control authority is maintained. Ultimately, this approach markedly curtails in-flight angular excursions and ensuring consistent upright landing configurations, offering a highly practical and control-efficient solution for traversing irregular extraterrestrial surfaces.
\end{abstract}


\section{INTRODUCTION}

Robotic exploration of low-gravity celestial bodies such as the Moon, asteroids, and planetary moons requires locomotion strategies fundamentally different from terrestrial paradigms. Early work by Fiorini et al. \cite{fiorini1999hopping} demonstrated that hopping robots can outperform wheeled and rocket-based mobility concepts in reduced gravity, showing that ballistic locomotion offers favorable energy efficiency derived from Apollo mission data. Their spring-driven egg-shaped hopper used an eccentric internal mass for directional control, highlighting the mechanical simplicity achievable for space deployment. However, attitude regulation was limited to pre-takeoff alignment, and no active in-flight stabilization was available during ballistic motion.

More recently, SpaceHopper \cite{spiridonov2024spacehopper} advanced low-gravity hopping using a three-legged architecture capable of mid-air inertial re-orientation through coordinated leg motion. Instead of reaction wheels, SpaceHopper redistributes angular momentum via multi-degree-of-freedom leg actuation and deep reinforcement learning (DRL) policies. Simulations under Ceres gravity ($0.029g$) demonstrated reliable 6 m ballistic jumps and upright reorientation, with partial hardware validation. While highly capable, this approach increases mechanical complexity and relies heavily on data-driven control.

\begin{figure}[t]
    \centering
    \includegraphics[width=\linewidth]{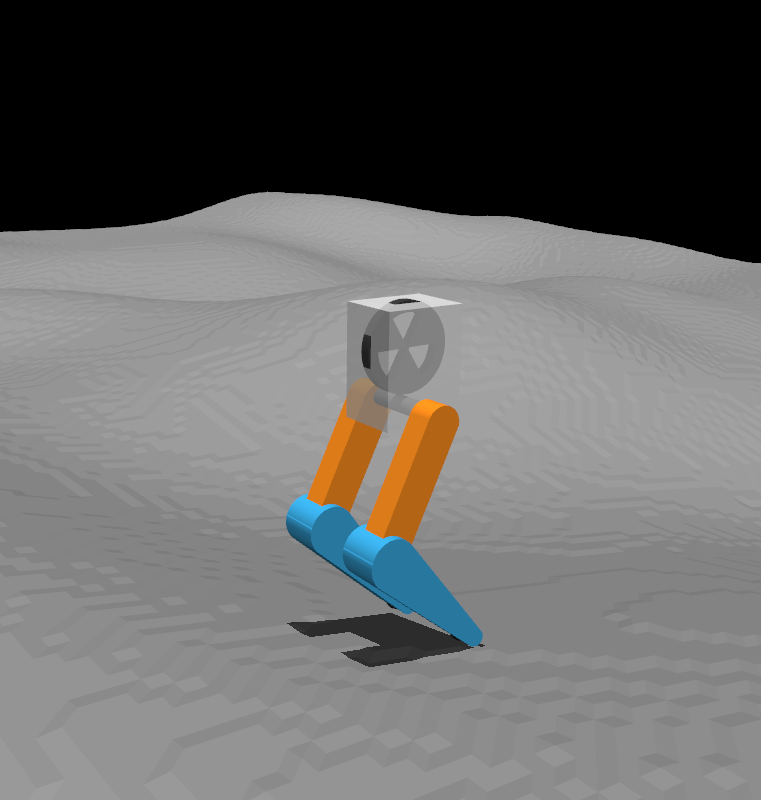}
    \caption{Overview of the reaction-wheel based bipedal hopper operating on procedurally generated low-gravity terrain.}
    \label{fig:mars_hopper}
\end{figure}

These two paradigms—actuator-minimal spring hoppers without flight stabilization \cite{fiorini1999hopping} and multi-legged DRL-based reorientation systems \cite{spiridonov2024spacehopper}—represent opposite ends of the design spectrum. The trade-off between mechanical simplicity, control tractability, and in-flight attitude authority remains insufficiently explored, particularly for reduced-degree-of-freedom systems intended for mass- and power-constrained planetary missions.

Reduced gravity significantly alters locomotion dynamics. Reduced ground reaction forces weaken traction-based mobility, while extended flight phases amplify rotational drift caused by asymmetric takeoff impulses or terrain irregularities. Studies on walking and jumping robots under reduced gravity conditions confirm that terrestrial control strategies degrade when gravitational acceleration changes \cite{omer2014lowgravity,sripada2018jumping}. Moreover, hybrid locomotion systems that alternate between stance and flight require coordinated handling of impact dynamics and control constraints \cite{lu2025runjump}. During ballistic flight, no external torques are available; therefore, orientation can only be regulated through internal momentum redistribution.

Inertial reorientation through internal appendages has been extensively studied in robotics. Theoretical foundations show that any internal mass motion—tails, limbs, or wheels—can induce body rotation via angular momentum conservation \cite{libby2016inertial}. A key metric is the appendage effectiveness,which is  determined by the inertia ratio between the appendage and the body. Tails often provide high moment-of-inertia efficiency but are limited by finite range of motion and potential self-collision. Reaction wheels, in contrast, offer unlimited rotational range and compact integration at the expense of lower inertia efficiency \cite{libby2016inertial}.

Terrestrial robots have adopted both approaches. Tail-based systems enable planar and spatial reorientation but may face actuation and collision constraints \cite{chu2023quadratic}. Reaction wheels have been successfully used for balance enhancement and mid-air stabilization in legged robots \cite{lee2023reactionwheel,kim2024jumping}. Hybrid appendage systems combining tails and wheels have also been proposed to exploit complementary advantages \cite{chu2023quadratic}. More abstract formulations demonstrate that legs themselves can function as reaction masses for body reorientation when external contacts are absent \cite{balasubramanian2003legless}. In humanoid systems, explicit angular momentum regulation has been shown to play a critical role in disturbance rejection and reactive stabilization \cite{yun2011momentum,nenchev2021momentum}.

In the context of hopping systems, stabilization has been explored using reaction wheels mounted on minimal platforms \cite{ryadchikov2018stabilization}, but such efforts have primarily targeted terrestrial gravity conditions. For extraterrestrial applications, multi-modal robotic platforms integrating specialized mechanisms have also been proposed \cite{zhang2025lunar}, yet the specific problem of minimal-actuation ballistic stabilization in sustained low-gravity flight remains comparatively under-addressed.

This paper addresses this gap by proposing a reaction-wheel–stabilized bipedal hopping robot specifically designed for low-gravity environments such as the Moon. Unlike appendage-based or DRL-driven reorientation strategies \cite{spiridonov2024spacehopper,chu2023quadratic}, the proposed system employs a single internally mounted reaction wheel located at the torso center of mass to regulate pitch during ballistic flight. The robot is modeled as a planar gyrostat, capturing the coupled rotational dynamics between body and wheel. A reduced-order formulation isolates the angular momentum exchange governing mid-air stabilization. The biped design further enables independent leg adjustment at touchdown for better accommodation of uneven terrain.

The operational cycle consists of three hybrid phases: (i) impulsive takeoff via leg actuation, (ii) ballistic flight with closed-loop reaction-wheel attitude control, and (iii) touchdown absorption through impedance-controlled legs. In contrast to learning-based methods, a classical PID controller is designed for the flight phase to emphasize analytical transparency, low computational demand, and suitability for resource-constrained space missions.

The proposed architecture is evaluated in a MuJoCo simulation configured for reduced-gravity conditions representative of lunar and Martian environments. The main contributions of this work are:
\begin{itemize}
    \item A mechanically minimal bipedal hopping architecture integrating onboard momentum exchange for low-gravity ballistic locomotion.
    
    \item A reduced-order gyrostat-based dynamic model describing the coupled torso-reaction-wheel behavior during ballistic flight.
    
    \item A classical PID-based in-flight attitude controller tailored specifically to low-gravity ballistic phases.
    
    \item Validation in physics-based reduced-gravity simulation demonstrating stable landing performance over irregular terrain.
\end{itemize}

\section{System Overview}

\subsection{Mechanical Configuration}

The robot body consists of a rigid torso connected to two symmetric articulated legs, each incorporating knee and ankle pitch joints. The reaction wheel is concentrically mounted at the torso center of mass to decouple its operation from translational dynamics. The complete actuator set comprises five drives: (1) left knee actuator, (2) left ankle actuator, (3) right knee actuator, (4) right ankle actuator, and (5) reaction wheel motor. This configuration forms an underactuated knife-edge biped, posing significant control challenges and limited actuation authority.
During the ballistic phase, the leg actuators are rendered mostly passive (with only minimal lift-off and landing adjustments), while the reaction wheel motor remains actively controlled, effectively reducing the system to a predominantly two-state rotational subsystem.

\subsection{Phase Transitions and Hybrid Nature}

The hopping cycle exhibits hybrid dynamics, involving discrete mode switches between ground contact and free flight. The transition criterion is based on the normal ground reaction force:
\begin{equation}
F_n = 0
\end{equation}
When $F_n \leq 0$, the system enters ballistic flight. Accurate characterization of these transitions requires hybrid modeling formulations 
\cite{chen2013hopping}

\section{Dynamic Modeling}

\subsection{Modeling Assumptions}
The system is modeled as a planar floating-base bipedal hopper operating in the sagittal plane. Each leg consists of two revolute joints (hip and knee). The torso is equipped with an internal reaction wheel aligned with the pitch axis. Ground contact occurs at the foot and is modeled as a holonomic constraint during stance.

\begin{figure}[t]
    \centering
    \includegraphics[width=0.75\linewidth]{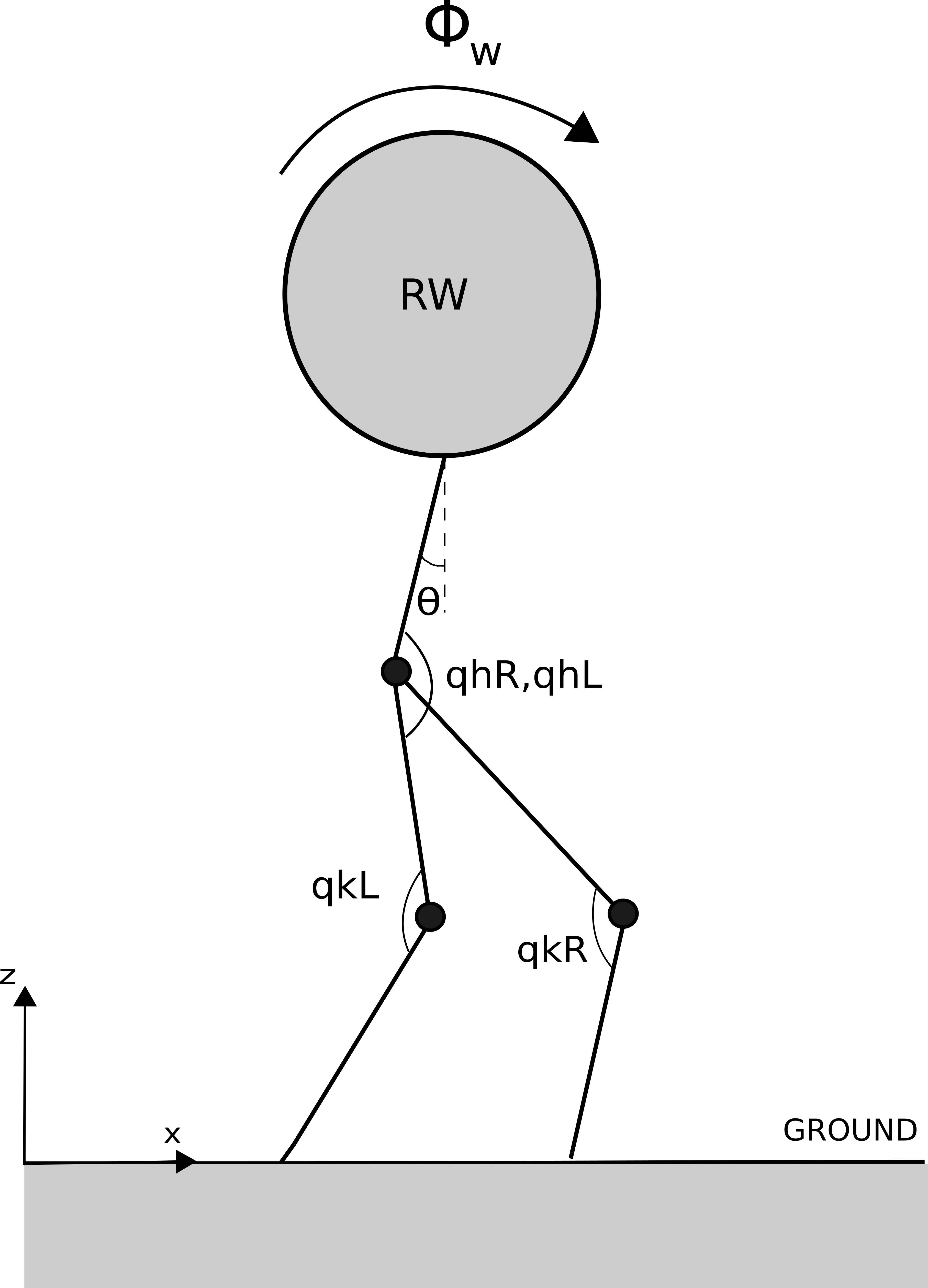}
    \caption{Planar dynamic model of the reaction-wheel hopper illustrating the torso-mounted reaction wheel (RW), wheel rotation $\Phi_w$, hip and knee joint angles ($q_{hL}, q_{hR}, q_{kL}, q_{kR}$), and the ground reference frame used for system modeling.}
    \label{fig:mars_hopper}
\end{figure}

\subsection{Generalized Coordinates}
The generalized coordinate vector for the planar model is defined as:
\begin{equation}
q = \begin{bmatrix} x & z & \theta & \phi_w & q_{hL} & q_{kL} & q_{hR} & q_{kR} \end{bmatrix}^T
\end{equation}
where $x,z$ are the torso Cartesian coordinates, $\theta$ is the torso pitch angle, $\phi_w$ is the reaction wheel angle relative to the torso, and $q_{h*},q_{k*}$ denote the hip and knee angles.

\subsection{Lagrangian Dynamics}
The kinetic energy $T$ includes contributions from the torso, reaction wheel, and both legs:
\begin{equation}
T = \frac{1}{2} m_b(\dot{x}^2+\dot{z}^2) + \frac{1}{2} I_b \dot{\theta}^2 + \frac{1}{2} I_w(\dot{\theta}+\dot{\phi}_w)^2 + T_{\text{legs}}(q,\dot{q})
\end{equation}
where $T_{\text{legs}}$ is the kinetic energy of the two legs, which can be derived from standard kinematic expressions but is omitted here for brevity. The potential energy is
\begin{equation}
V = m_b g z + V_{\text{legs}}(q).
\end{equation}
Forming the Lagrangian $\mathcal{L}=T-V$ and applying the Euler-Lagrange equations yields the manipulator form:
\begin{equation}
M(q)\ddot{q} + C(q,\dot{q})\dot{q} + G(q) = B\tau + J_c^T F_c,
\label{eq:manipulator}
\end{equation}
where $M(q)$ is the inertia matrix, $C(q,\dot{q})$ contains Coriolis and centrifugal terms, $G(q)$ is the gravity vector, $\tau$ is the actuator torque vector, $B$ maps actuators to generalized forces, and $J_c^T F_c$ represents contact forces during stance.

\subsection{Actuation Structure}
The actuator torque vector is
\begin{equation}
\tau = \begin{bmatrix} \tau_w & \tau_{hL} & \tau_{kL} & \tau_{hR} & \tau_{kR} \end{bmatrix}^T.
\end{equation}
The floating base (coordinates $x,z,\theta$) is unactuated, so $B$ has rank 5 while $\dim(q)=8$, rendering the system underactuated.

\subsection{Reaction Wheel Coupling}
The total angular momentum about the torso pitch axis is
\begin{equation}
H = I_b \dot{\theta} + I_w(\dot{\theta}+\dot{\phi}_w).
\end{equation}
Differentiating and using $\tau_w = I_w \ddot{\phi}_w$ gives
\begin{equation}
(I_b+I_w)\ddot{\theta} = \tau_{\text{ext}} - \tau_w,
\label{eq:reaction wheel_coupling}
\end{equation}
where $\tau_{\text{ext}}$ is the external torque about the pitch axis (zero during flight). Equation~\eqref{eq:reaction wheel_coupling} shows that reaction wheel torque directly affects torso angular acceleration, enabling internal momentum exchange.

\subsection{Hybrid Dynamics}
The system evolves through two distinct phases:
\begin{align}
\text{Flight:} \quad & M(q)\ddot{q} + C(q,\dot{q})\dot{q} + G(q) = B\tau, \\
\text{Stance:} \quad & M(q)\ddot{q} + C(q,\dot{q})\dot{q} + G(q) = B\tau + J_c^T F_c.
\end{align}
Transitions between phases are governed by ground contact events and are modeled with impulsive impact maps.

cit\section{Control Design}
\subsection{State-Space Formulation}
Define the state vector as:
\[
x = \begin{bmatrix} \theta \\ \omega \end{bmatrix}
\]

The continuous-time state-space representation is:
\[
\dot{x} = Ax + Bu
\]
where
\[
A = \begin{bmatrix} 0 & 1 \\ 0 & 0 \end{bmatrix}, \quad
B = \begin{bmatrix} 0 \\ -\frac{1}{I_b + I_w} \end{bmatrix}, \quad
u = \tau_w
\]

\subsection{PID Control Formulation}

The control objective is to regulate torso pitch angle to zero during the ballistic phase. A classical PID controller is implemented based on torso angle error:

\[
e(t) = \theta(t)
\]

The control torque applied to the reaction wheel is defined as:

\begin{equation}
\tau_w = -K_p e(t) - K_d \dot{e}(t) - K_i \int_0^t e(\tau) d\tau
\end{equation}

where:
\begin{itemize}
    \item $K_p$ is the proportional gain
    \item $K_d$ is the derivative gain
    \item $K_i$ is the integral gain
\end{itemize}


\subsection{Reaction Wheel Handling}

Reaction wheel saturation arises when the spin rate reaches its mechanical limit:
\begin{equation}
|\omega_w| \geq \omega_{max}
\end{equation}
A baseline desaturation scheme is applied during the stance phase by injecting a counter-torque through ground contact forces. More sophisticated formulations, including optimal desaturation over the stance interval, remain a subject for future investigation.

\section{Simulation Analysis }




The terrain is modeled using a procedurally generated heightfield representing a cratered lunar surface. A synthetic elevation map is created in Python by superimposing a radially symmetric crater profile and exported as a grayscale image, which is then imported into MuJoCo as a heightfield asset.
\subsection{Simulation Parameters}

The physical, actuator, and simulation parameters used in the MuJoCo environment 
are summarized in Table~\ref{tab:sim_params}.

\begin{table}[h]
\centering
\caption{Simulation and Physical Parameters}
\label{tab:sim_params}
\begin{tabular}{l c}
\hline
\multicolumn{2}{c}{\textit{Simulation Settings}} \\
\hline
Gravity $g$ & $1.625$ m/s$^2$ (Moon) \\
Timestep $\Delta t$ & $0.001$ s \\
Control frequency & $1$ kHz \\
\hline
\multicolumn{2}{c}{\textit{Body Parameters}} \\
\hline
Torso mass & $4.0$ kg \\
Reaction wheel mass & $3.0$ kg \\
Total system mass & $8.2$ kg \\
\hline
\multicolumn{2}{c}{\textit{Leg Parameters (per leg)}} \\
\hline
Upper leg mass & $0.4$ kg \\
Lower leg mass & $0.2$ kg \\
Upper leg length & $0.5$ m \\
Lower leg length & $0.5$ m \\
Hip joint range & $[-120^\circ, 120^\circ]$ \\
Knee joint range & $[-160^\circ, 0^\circ]$ \\
\hline
\multicolumn{2}{c}{\textit{Contact}} \\
\hline
Foot contact sensor radius & $0.04$ m \\
\hline
\end{tabular}
\end{table}

The selected actuator and inertia parameters were chosen to ensure stable numerical behavior and sufficient control authority within the simulation environment. Body pitch angle measurements are obtained from the IMU sensor, while ground contact is detected using foot-mounted contact sensors, both within the MuJoCo environment.

\subsection{Performance Metrics}

System behavior is quantified through the following criteria:
\begin{itemize}
    \item Peak mid-air angular deviation
    \item Root mean square (RMS) torso angular velocity
  
\end{itemize}

\section{Results and Discussion}

The proposed reaction wheel (RW) stabilization framework was evaluated across a continuous, multi-hop locomotion sequence over uneven terrain in a simulated low-gravity environment ($g = 1.625 \text{ m/s}^2$). System behavior is quantified through continuous kinematic tracking and discrete hop-by-hop performance metrics to assess mid-air attitude control, energy stability, safety constraints, and actuator limits.

\subsection{Hopping Kinematics and Terrain Adaptation}

\begin{figure}[htbp]
    \centering
    \begin{subfigure}{0.48\columnwidth}
        \includegraphics[width=\columnwidth]{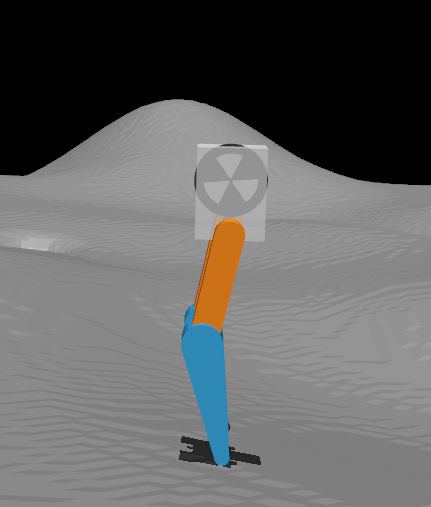}
        \caption{$t_1$: Touchdown}
    \end{subfigure}
    \hfill
    \begin{subfigure}{0.48\columnwidth}
        \includegraphics[width=\columnwidth]{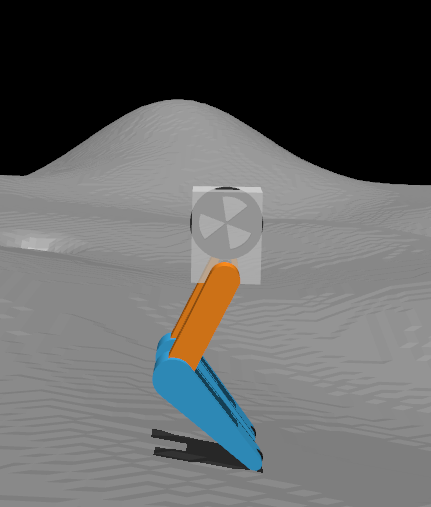}
        \caption{$t_2$: Crouch and Lift-Off}
    \end{subfigure}
  
    \vskip\baselineskip
    
    \begin{subfigure}{0.48\columnwidth}
        \includegraphics[width=\columnwidth]{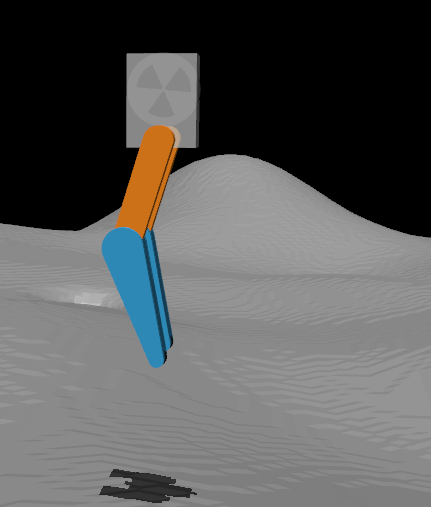}
        \caption{$t_3$: Mid-Air Stabilization}
    \end{subfigure}
    \hfill
    \begin{subfigure}{0.48\columnwidth}
        \includegraphics[width=\columnwidth]{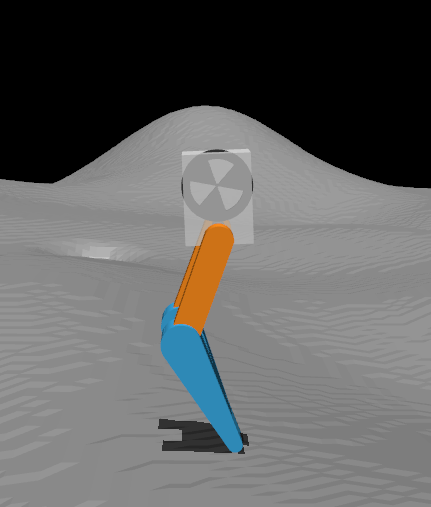}
        \caption{$t_4$: Landing}
    \end{subfigure}
    
    \caption{Representative time instants of a single hopping cycle illustrating touchdown, propulsion, mid-air stabilization, and landing phases.}
    \label{fig:hopping_sequence}
\end{figure}

Figure \ref{fig:hopping_sequence} illustrates the representative physical states of the bipedal hopper during a single complete hopping cycle. The sequence highlights the successful execution of energy absorption during touchdown ($t_1$), potential energy storage in the deep crouch phase and propulsion ($t_2$), ballistic flight and RW-driven attitude correction ($t_3$), and preparation for the subsequent landing ($t_4$). The developed system achieves a horizontal hopping range of approximately 0.8 m per cycle.

\begin{figure}[htbp]
    \centering
    \begin{subfigure}{0.8\columnwidth}
        \includegraphics[width=\columnwidth]{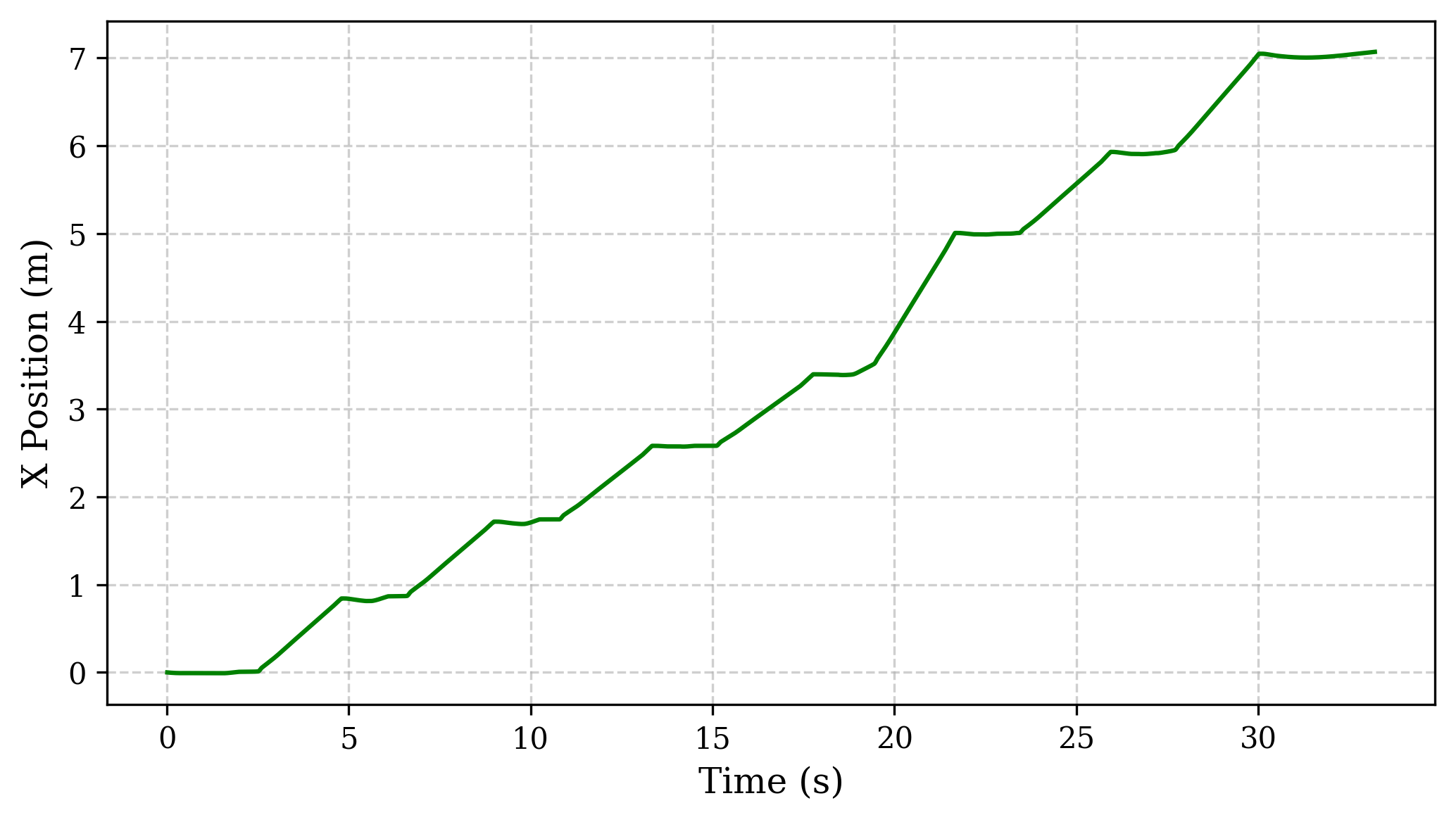}
        \caption{Forward (X) Translation}
    \end{subfigure}
    
    \vspace{\baselineskip}
    
    \begin{subfigure}{0.8\columnwidth}
        \includegraphics[width=\columnwidth]{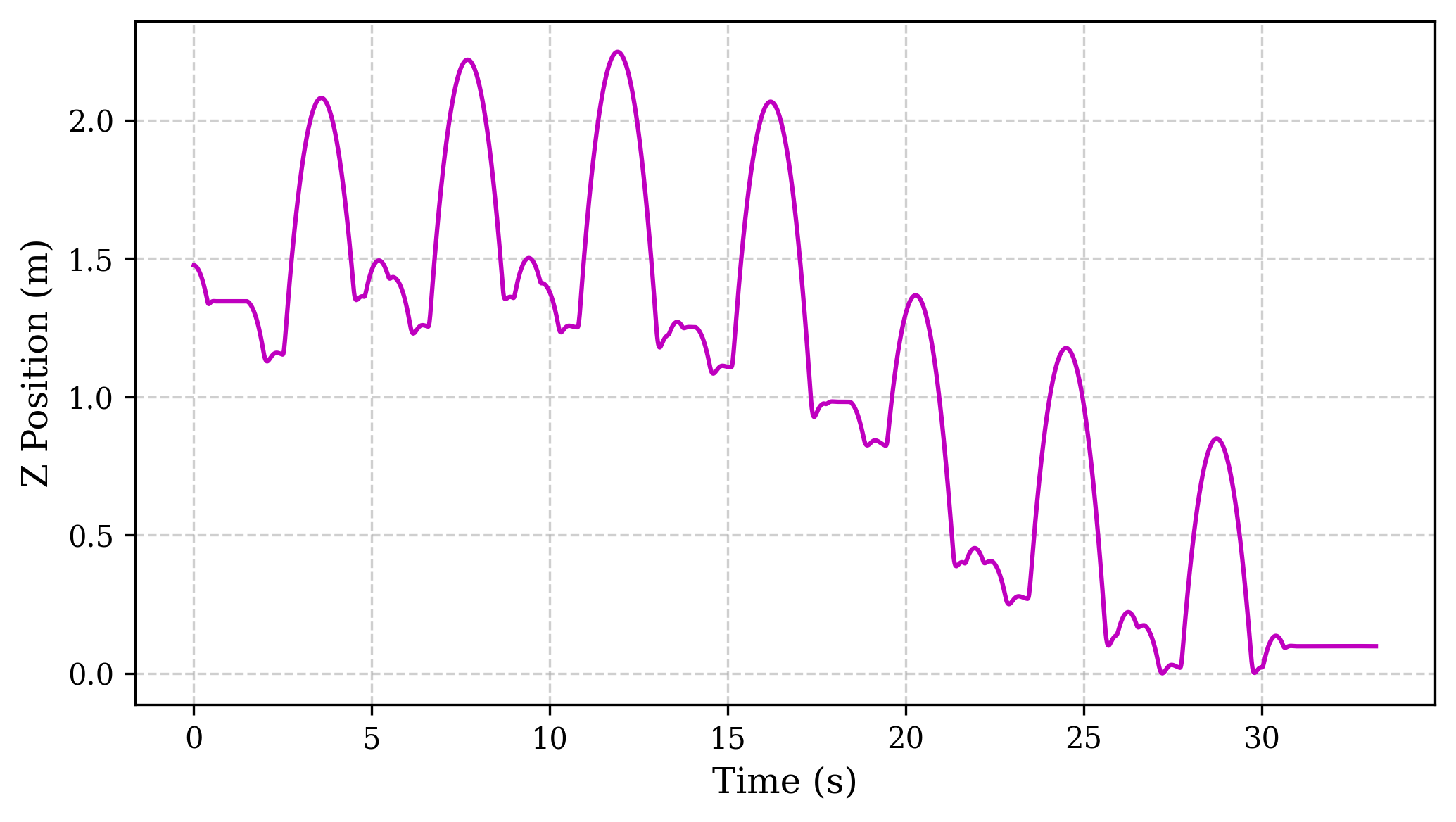}
        \caption{Vertical (Z) Position}
    \end{subfigure}
    
    \caption{Continuous spatial trajectories across successive hops. The changing vertical baseline indicates successful adaptation to the uneven lunar heightmap, while forward progression remains consistent.}
    \label{fig:continuous_translation}
\end{figure}

The macro-level mobility of the robot over the 7-hop sequence is validated by its continuous spatial trajectories. As shown in Figure \ref{fig:continuous_translation}, the forward (X) and vertical (Z) translation profiles confirm that the state machine successfully adapts to the uneven heightmap. The consistent "staircase" pattern in the forward translation and the steadily rising and falling baseline in the hopping height prove that the robot maintains its rhythm without stalling, even as the terrain elevation gradually changes. The smaller intermediate oscillations in the Z-axis correspond to the crouch phase, where knee flexion under low-gravity conditions induces a slight, unintended lift prior to the main push-off.

\subsection{Mid-Air Attitude Stabilization and System Energy}

During the ballistic flight phases ($t_4$), the aggressive backward sweep of the leg induces significant pitching moments on the torso. In the absence of RW control, prior tests indicated that initial pitch perturbations of $15^\circ$ to $20^\circ$ routinely escalated into uncontrolled tumbling. 

\begin{figure}[htbp]
    \centering
    \begin{subfigure}{0.8\columnwidth}
        \includegraphics[width=\columnwidth]{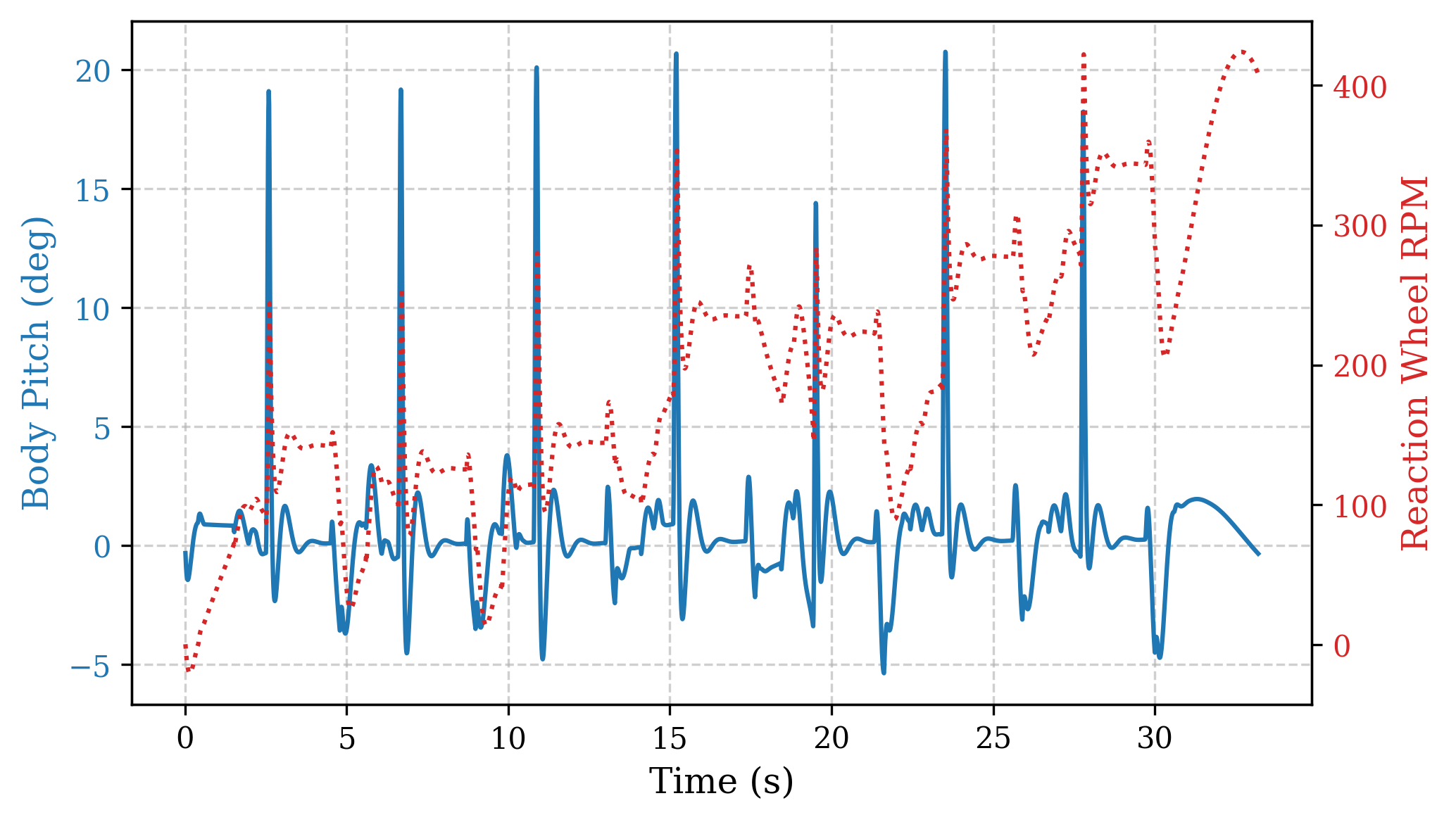}
        \caption{Body Pitch vs. RW RPM}
    \end{subfigure}
    
    \vspace{\baselineskip}
    
    \begin{subfigure}{0.8\columnwidth}
        \includegraphics[width=\columnwidth]{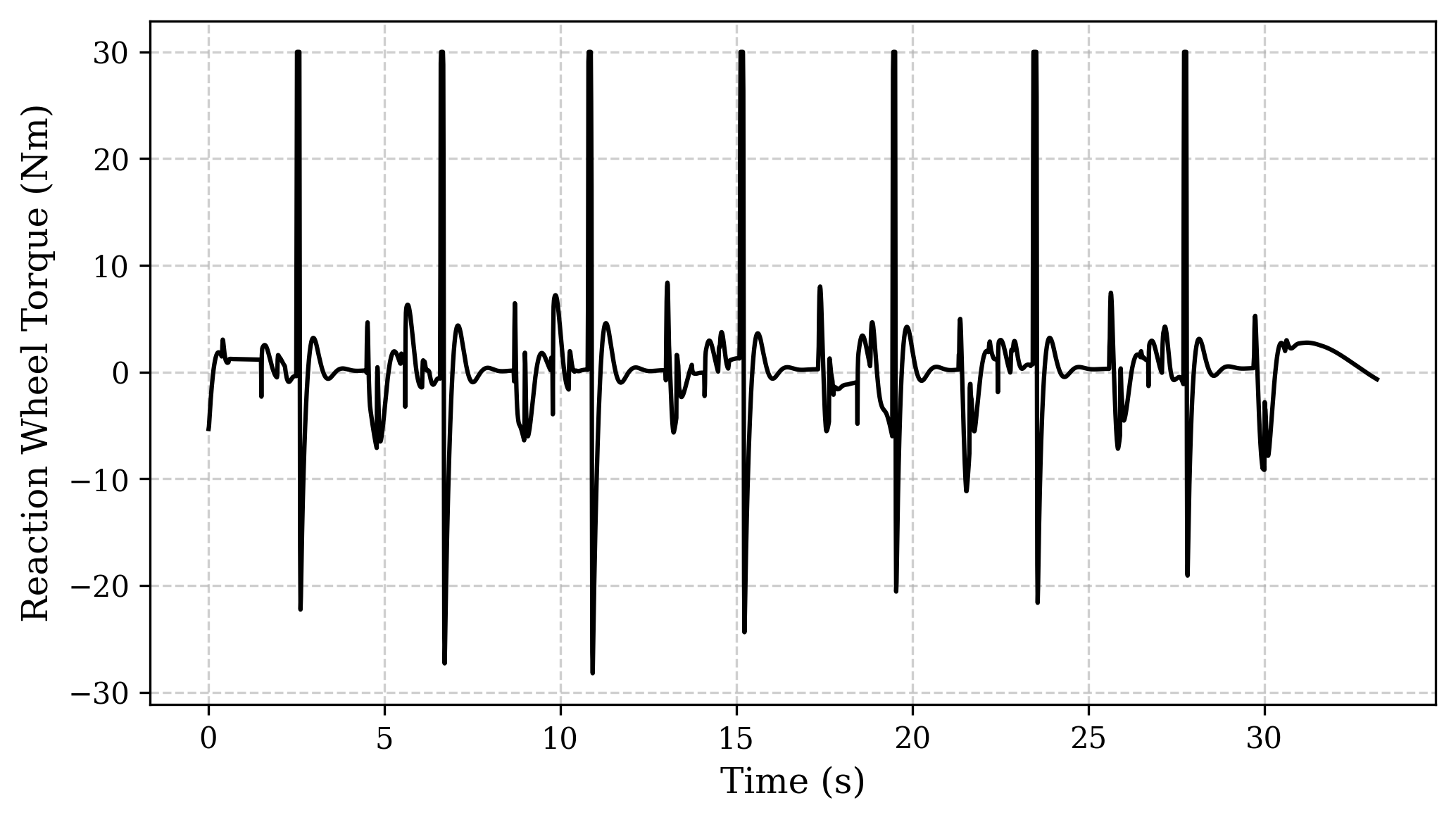}
        \caption{Control Effort (Torque)}
    \end{subfigure}
    
    \caption{Reaction wheel dynamic response. The wheel accelerates dynamically to absorb liftoff-induced angular momentum, with torque impulses aligning strictly with physical push-off and landing disturbances.}
    \label{fig:control_response}
\end{figure}

With the active controller engaged, the reaction wheel rapidly spins up to absorb this angular momentum (Figure \ref{fig:control_response}a). The corresponding torque profile (Figure \ref{fig:control_response}b) exhibits sharp, transient impulses precisely synchronized with liftoff and touchdown events. These impulses successfully suppress pitch disturbances, forcibly driving the torso back toward zero pitch mid-air.

\begin{figure}[htbp]
    \centering
    \begin{subfigure}{0.8\columnwidth}
        \includegraphics[width=\columnwidth]{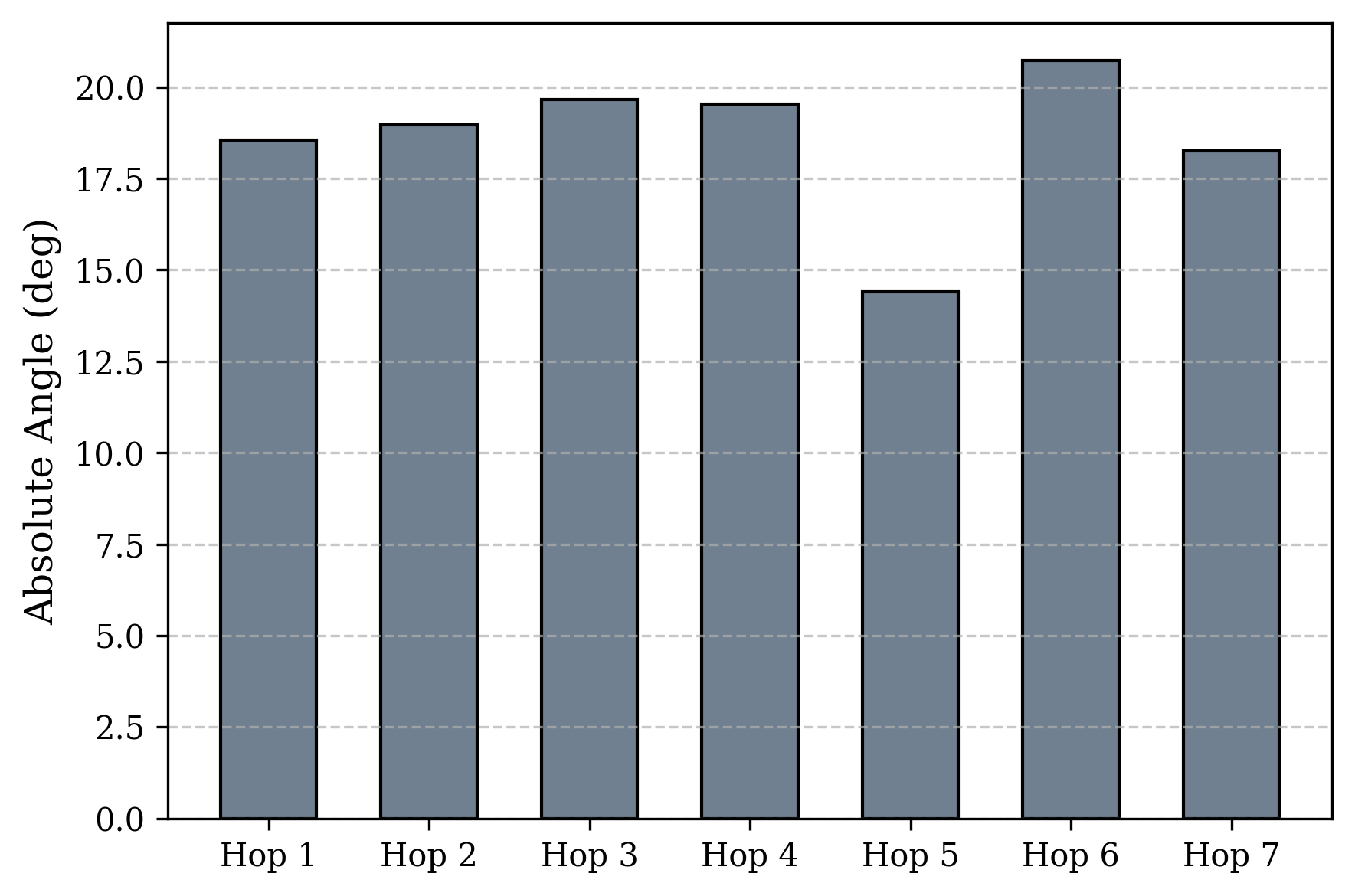}
        \caption{Peak Mid-Air Pitch Deviation}
    \end{subfigure}
    
    \vspace{\baselineskip}
    
    \begin{subfigure}{0.8\columnwidth}
        \includegraphics[width=\columnwidth]{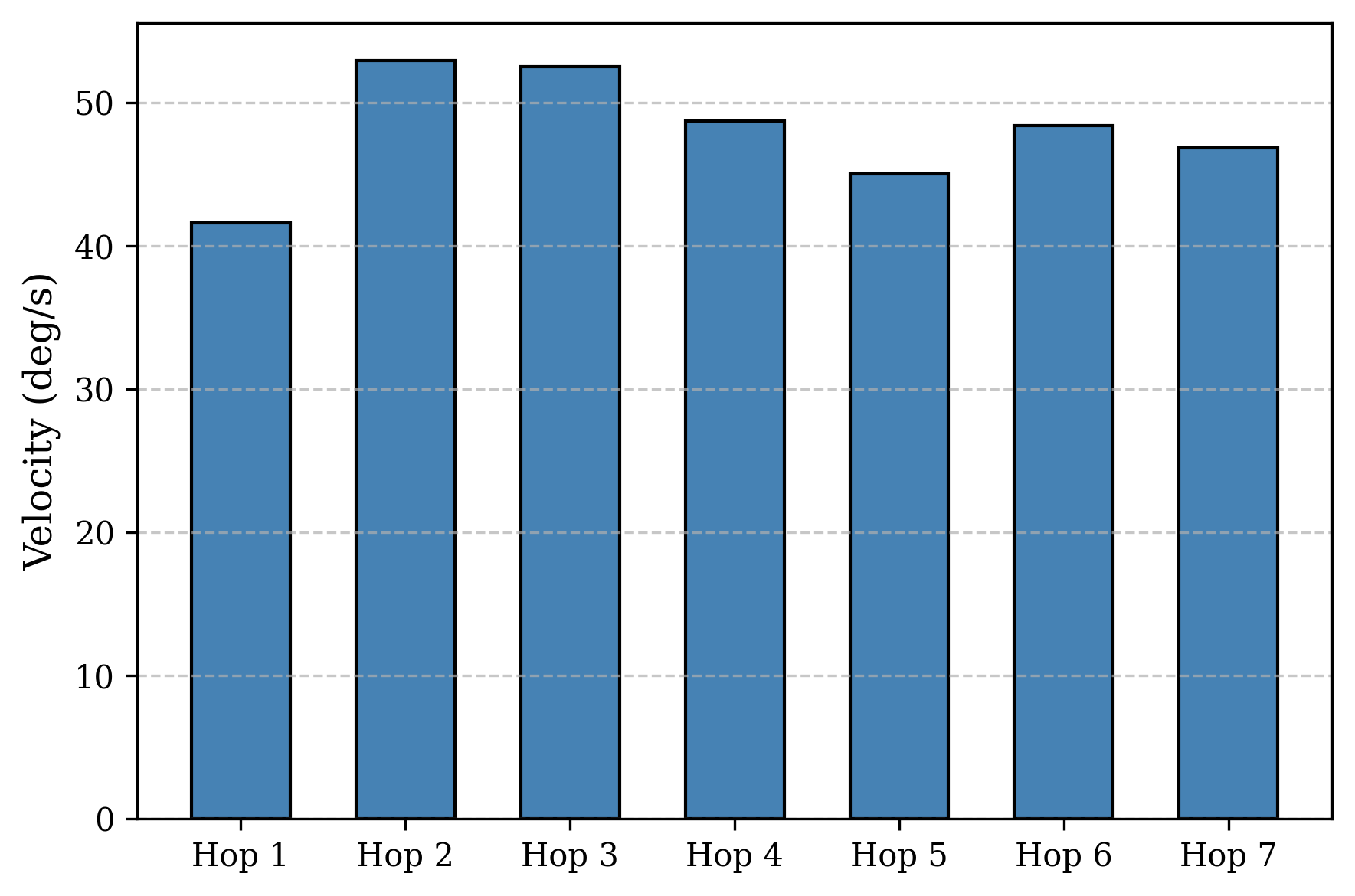}
        \caption{Torso RMS Angular Velocity}
    \end{subfigure}
    
    \caption{Hop-by-hop performance metrics evaluating system stability and landing safety.}
    \label{fig:metrics_summary}
\end{figure}

This continuous behavior is mathematically validated by the discrete performance metrics in Figure \ref{fig:metrics_summary}. As quantified in Figure \ref{fig:metrics_summary}(a), the peak mid-air angular deviation was consistently bounded between $15^\circ$ and $20^\circ$ following the leg sweep, demonstrating the controller's effectiveness in strictly capping momentum accumulation. Furthermore, data tracking indicated that the root mean square (RMS) torso angular velocity remained stable between $40^\circ/\text{s}$ and $55^\circ/\text{s}$ across all hops. This consistency indicates the system has reached a stable limit cycle, preventing high-frequency kinetic energy from compounding across sequential leaps.

\subsection{Touchdown Accuracy and Landing Stability}

Safe biped locomotion relies heavily on the orientation of the center of mass relative to the contact point immediately prior to touchdown. If the torso is heavily pitched upon impact, ground-reaction forces will generate an unrecoverable torque multiplier, causing loss of balance.

Because the reaction wheel stabilization system successfully zeros the mid-air pitch, the torso consistently converges toward an upright posture before ground contact is re-established. This near-perfect touchdown configuration drastically reduces asymmetric landing impacts and guarantees the repeatability of the stance phase.

\subsection{Actuator Saturation and Hardware Feasibility}

While extended ballistic intervals in low-gravity environments afford the controller additional time for momentum exchange, prolonged exposure to these control demands increases the likelihood of actuator saturation. If the reaction wheel hits its maximum torque or RPM limits, the system loses control authority and becomes uncontrollable.

To validate the real-world hardware feasibility of this design, cumulative reaction wheel saturation time (defined as time spent operating at $\ge 98\%$ of the $29.5 \text{ Nm}$ torque limit) was tracked per hop. Saturation time was severely restricted, registering at less than $0.9$ seconds per cycle. This confirms that the selected $30:1$ gear ratio and motor specifications are appropriately sized for the torso mass. The controller successfully rejects peak disturbances while maintaining adequate control authority, underscoring the viability of this mechanical architecture for planetary exploration platforms.

\section{Conclusion}

In this work, a reaction-wheel–stabilized hopping robot was modeled and simulated in a reduced-gravity Moon-based environment using a physics-based MuJoCo framework. A structured contact-driven state machine was developed to regulate landing, stance stabilization, crouch loading, push-off, and aerial phases. Attitude stabilization in the sagittal plane was achieved using reaction wheel control, enabling controlled hopping cycles over uneven height-field terrain. The system demonstrated repeatable ballistic motion and stable landing behavior under reduced gravity conditions. 

Future research directions include hardware fabrication and experimental validation on a physical prototype, extension to full three-dimensional attitude stabilization including yaw dynamics, development of proactive reaction wheel desaturation strategies, optimization of the hopping sequence for improved stability and repeatability, formulation of energy-optimal control policies for the hopping cycle, and integration of terrain-perception-based MPC or iLQR control for enhanced efficiency and predictive landing performance. In addition, reinforcement learning (RL) based control is to be explored for improving robustness and adaptive behavior across varying terrain conditions.

\end{document}